\documentclass[%
 aip,
 amsmath,amssymb,
 reprint,%
]{revtex4-1}

\usepackage[english]{babel}

\usepackage[letterpaper,top=2cm,bottom=2cm,left=3cm,right=3cm,marginparwidth=1.75cm]{geometry}

\usepackage{centernot}
\usepackage{graphicx}
\usepackage{comment}
\newtheorem{theorem}{Theorem}[section]
\usepackage[colorlinks=true, allcolors=blue]{hyperref}
\usepackage{subfig}
\usepackage[]{xcolor}

\begin{document}

\title{Time-shift selection for reservoir computing using a rank-revealing QR algorithm
}

\author{Joseph D. Hart}
\email{joseph.hart@nrl.navy.mil}
\affiliation{US Naval Research Laboratory, Washington, DC 20375}

\author{Francesco Sorrentino}
\affiliation{Department of Mechanical Engineering, University of New Mexico, Albuquerque, NM 87131}

\author{Thomas L. Carroll}
\affiliation{US Naval Research Laboratory,  Washington, DC 20375}

\date{\today}

\begin{abstract}
Reservoir computing, a recurrent neural network paradigm in which only the output layer is trained, has demonstrated remarkable performance on tasks such as prediction and control of nonlinear systems. Recently, it was demonstrated that adding time-shifts to the signals generated by a reservoir can provide large improvements in performance accuracy. In this work, we present a technique to choose the time-shifts by maximizing the rank of the reservoir matrix using a rank-revealing QR algorithm. This technique, which is not task dependent, does not require a model of the system, and therefore is directly applicable to analog hardware reservoir computers. We demonstrate our time-shift selection technique on two types of reservoir computer: one based on an opto-electronic oscillator and the traditional recurrent network with a $tanh$ activation function. We find that our technique provides improved accuracy over random time-shift selection in essentially all cases.

\end{abstract}

\maketitle

\definecolor{cerulean}{rgb}{.121,0.465,0.703}

\textbf{A reservoir computer is a type of recurrent neural network in which only the output layer is trained \cite{maass2002real,jaeger2004harnessing} that has displayed impressive performance at dynamical tasks such as prediction \cite{jaeger2004harnessing,pathak2018model,gauthier2021predicting} and control \cite{canaday2021model} of nonlinear systems. It is often desirable to have a small number of nodes in the reservoir network, as this can improve the speed, cost, size, and power consumption of the reservoir computer. However, the reservoir must still have a sufficient number of nodes to perform the desired task. A recent technique--augmenting the reservoir matrix with time-shifted versions of the reservoir states \cite{del2021reservoir, carroll2022time}--has been shown to enable good performance for reservoirs as small as five nodes \cite{carroll2022time}. In this work, we present a method to determine the time-shifts by selecting the time-shifts 
that maximize the rank of the reservoir matrix. Importantly, our
technique does not require a model of the system, and therefore is applicable for both software- and hardware-based reservoir computers. We find that our technique provides
up to 70\% improvement in accuracy over random time-shift selection.}

\section{Introduction}
Reservoir computing is a machine learning modality that is designed to be easy to train \cite{maass2002real,jaeger2004harnessing}. The main component of a reservoir computer (RC) is a dynamical system (``the reservoir'') that provides a nonlinear mapping of input signals into a higher dimensional space. Often, the reservoir is a random recurrent neural network; however, a variety of dynamical systems have also been shown to be effective reservoirs. 

Importantly, the reservoir itself is not trained. The only training that is performed is on the linear read-out layer of the reservoir state variables. The benefit of this training, which is typically done by ridge regression, is that it can be done at low computational expense and does not rely on a model of the reservoir. This latter feature has resulted in the utilization of a wide variety of analog hardware as the reservoir \cite{tanaka2019recent,van2017advances,du2017reservoir}. Hardware RCs show promise for high speed, low size and power computation \cite{larger2017high}.  

It is often desirable to have a small number of nodes in a reservoir network, as this can improve the speed, cost, size, and power consumption of hardware RCs \cite{sugano2019reservoir}. However, the reservoir must still have a sufficient number of nodes to perform the desired task. As a result of these competing demands, reservoir augmentation methods have been developed to increase the number of time series produced by the reservoir without increasing the number of physical nodes \cite{jaurigue2021reservoir,carroll2021adding,takano2018,harkhoe2020,sunada2021,marquez2019takens}. 

In this work, we consider a recently-developed technique: augmenting the reservoir matrix with time-shifted versions of the reservoir states \cite{del2021reservoir, carroll2022time}. This time-shifting technique has been shown to enable good performance for reservoirs as small as five nodes \cite{carroll2022time}. The crucial question addressed here is how to choose the time-shifts.

Del Frate et al. chose the node/time-shift combinations randomly \cite{del2021reservoir}, and Carroll et al. chose ordered time-shifts \cite{carroll2022time}. In this work, we present a method to determine the node/time-shift combinations by first creating the matrix $\mathbf{\Omega^{(2)}}$ of all nodes and time-shifts up to $\tau_{max}$, then retaining only the most linearly independent $M_{red}$ node/time-shift combinations. These combinations are identified using a rank-revealing QR algorithm, which is based on the standard QR factorization \cite{olver2006applied}. A significant benefit of this method is that it does not require a model of the system, and therefore is directly applicable to analog RCs, for which a model may not exist. Additionally, our method is not task dependent. There are times when one may want to set up a reservoir in a task-independent way; for example, one may want to perform more than one task simultaneously (e.g., observer and prediction tasks, as we do here). We find that our choice of time-shifts often result in significantly enhanced performance compared with random time-shifts in both opto-electronic and digital $tanh$ RCs.

\section{Reservoir computing}
\label{rescomp}

The most common type of reservoir is a recurrent neural network, and so we adopt this notation. In this work, the state variables of the reservoir are called ``nodes.''

A standard model for a RC is a network of $M$ coupled nonlinear maps given by 
\begin{equation}
    \mathbf{\chi}[n+1]=F(\mathbf{A\chi}[n] + \textbf{W}^{in}s[n]),
\end{equation}
where $\mathbf{\chi}[n]$ is an $M\times1$ vector describing the state of each network node $\chi_i$ at time $n$, $F(q)\equiv (F(q_1),F(q_2),...,F(q_M))$ is a nonlinear function applied element-wise to the $M\times 1$ vector $q$, \textbf{A} is the adjacency matrix of the network, $\textbf{W}^{in}$ is an $M\times1$ vector of input weights, and $s[n]$ is the input signal that drives the reservoir network.

In the training stage, the RC is driven with the input signal $s[n]$ with $0\leq n<T$ to produce the $M$ RC output signals $\chi_i[n]$. The $T\times M$ reservoir output matrix $\Omega$ is constructed from the reservoir signals as
\begin{equation}
\label{omegaeq}
\Omega_{n,j}=\chi_j[n]
\end{equation}

The reservoir has $M$ nodes and the input time series has $T$ points. The trained RC output $h[n]$ is obtained from $h[n]=\mathbf{\Omega}\mathbf{W}^{out}$, where $\mathbf{W}^{out}$ is an $M\times1$ vector of training coefficients obtained by minimizing the error between $h[n]$ and a training signal $g[n]$ via ridge regression \cite{tikhonov1977solutions}. 

In the testing stage, a new input signal $\tilde{s}[n]$ with corresponding test signal $\tilde{g}[n]$ is used to drive the reservoir, which produces $M$ output sequences, each of length $T_{test}$. A $T_{test}\times M$ reservoir matrix $\mathbf{\Omega_{test}}$ is formed similarly to $\mathbf{\Omega}$. The reservoir prediction $\tilde{h}[n]$ is given by $\tilde{h}[n]=\mathbf{\Omega W}^{out}$. The normalized root-mean-square testing error (NRMSE) $\Delta_{test}$ is computed as

\begin{equation}
\label{eq:training_error}
\Delta_{test}=\sqrt{\frac{1}{T_{test}}\sum_{n=1}^{T_{test}}\frac{(\tilde{g}[n]-\tilde{h}[n])^2}{(\tilde{g}[n]-<\tilde{g}>)^2}}, 
\end{equation}
where $<\cdot>$ indicates an average over time $T_{test}$.


\subsection{Reservoir rank and reservoir computing performance}

A key theme in this work is the relationship between reservoir covariance rank and reservoir computing performance, so we discuss this relationship here. We have previously shown that a RC's performance is strongly correlated with the reservoir's covariance rank \cite{carroll2019network,carroll2022time}. The covariance rank is commonly used in principal component analysis \cite{abdi2010principal} to determine the number of uncorrelated basis vectors required to represent a data matrix $\mathbf{\Omega}$, and is defined as the rank of the covariance matrix $\mathbf{\Sigma}=\mathbf{\Omega}^T\mathbf{\Omega}$.

The explanation for the correlation between reservoir performance and covariance rank is the following: 
The columns of a reservoir matrix with a large covariance rank span a large space. This improves the chances that the desired signal lies in the reservoir matrix column space, so a reservoir matrix with a large rank has a better chance to contain the required nonlinearities and memory to perform a generic task than does a reservoir matrix with a smaller rank. In other words, on average, a reservoir with a large rank will out-perform a reservoir with a smaller rank. Of course, this is not a guarantee: A specific task may require a specific nonlinearity that is not displayed by a reservoir with a large rank but is displayed by smaller rank reservoir. In this case, the smaller rank reservoir may indeed display higher performance. However, for an unknown task (or for a set of tasks that require different nonlinearities, such as the observer and prediction tasks considered here), a larger rank reservoir matrix is more likely to display better performance, as demonstrated in Ref. \cite{carroll2019network,carroll2022time}.

The reservoir covariance rank depends on the drive signal, but does not depend on the task to be performed. It also does not provide any information about reservoir consistency \cite{uchida2004consistency} or how to choose $\tau_{max}$ when using time-shifts.

\subsection{Augmenting reservoirs with time-shifts \label{sec:timeshifts}}


Consider a reservoir matrix (as in Eq. 2) with matrix elements $\Omega_{j,k}$.  We define $\Omega^{(2)}$ as the $(T-\tau_{max}) \times (\tau_{max}+1)M$ matrix in which each column is the time series of a specific node/time-shift combination for all time-shifts up to and including $\tau_{max}$:

\begin{equation}
\label{omega2}
\Omega^{(2)}_{j,k}=\Omega_{j+\tau_{max}-\lfloor k/M\rfloor,\mathrm{mod}(k,M)}
\end{equation}
where $\lfloor x\rfloor$ denotes the floor of $x$, $j\in\{0,...,T-\tau_{max}\}$, and $k\in\{0,1,...,M(\tau_{max}+1)-1\}$.
Using time shifts can be thought of (and implemented with minimal latency) as a digital filter with $\tau_{max}+1$ taps applied to each node \cite{carroll2021adding}, and the reservoir training as the optimization of that filter.

In this work, we use $\tau_{max}=10$ time steps, as done in Ref. \cite{carroll2022time}. Adding time-shifts to the reservoir matrix has been shown to be a simple yet effective way to improve RC performance that works by increasing the reservoir rank \cite{del2021reservoir,carroll2022time}.

\subsection{Input signals}
In this work, we characterize RCs by their ability to perform short-time prediction and observer tasks on two different chaotic systems: the Lorenz system and the R{\"o}ssler system. For the prediction task, we drive the RC with the $x$ variable of the Lorenz or R{\"o}ssler system, and train it to predict the next time step of the time series. For the observer task, we drive the RC with the $x$ variable of the chaotic system and use the RC to infer the $z$ variable. For each task, we use 8000 training steps and 7500 testing steps. 

The ordinary differential equations (including parameters) describing the Lorenz and R{\"o}ssler systems are the same as used in Ref. \cite{carroll2022time}. 
Both systems were sampled with unit time step.

\section{Rank-optimization of time-shifts using RRQR}\label{sec:rrqr}

It may often be desirable to use a subset of all time-shifts as this can lead to reduced computational expense in training and operating a RC. In these situations one must select the node/time-shift combinations to be used. One way to make this choice that depends on the numerical time-derivative of each node was presented in Ref. \cite{del2021reservoir}; that method is equivalent to using the full $\mathbf{\Omega^{(2)}}$ and setting $\tau_{max}=1.$

As stated previously, the reservoir covariance rank gives the number of linearly independent principal components in the reservoir matrix. Motivated by this relationship, in this section we develop a method to select the ``best'' node/time-shift combinations by determining the $M_{red}$ most linearly independent columns of the time-shifted reservoir matrix $\mathbf{\Omega^{(2)}}$. The time-shift/node combinations are optimal in this sense, and we refer to them as ``rank-optimal.'' The rank-optimal time-shifts are identified using a rank-revealing QR (RRQR) algorithm \cite{chan1987rank}.

In general, the RRQR algorithm is a method of extracting the minimal number of columns (node/time-shift combinations) that span the column space of a matrix (response space of a RC). However, many RCs with time-shifts are full rank or nearly full rank. We will show that the RRQR rank-optimization is still useful in these scenarios: It can be used to provide a ranking of the ``best'' node responses, and can therefore be used to select the ``best'' $M_{red}$ time-shift/node combinations. 

\subsection{QR decomposition with pivoting \label{sec:QR}}
We briefly summarize the QR decomposition with pivoting, following Ref. \cite{chan1987rank}. Any $T\times M$ matrix $\mathbf{B}$ with $T\ge M$ may be decomposed as 
\begin{equation}
    \mathbf{B\Pi}=\mathbf{QR},
\end{equation}
where $\mathbf{Q}$ is a $T\times T$ unitary matrix, $\mathbf{R}$ is a $T\times M$ upper triangular matrix, and $\mathbf{\Pi}$ is a $M\times M$ permutation matrix. $\mathbf{\Pi}$ is chosen such that the diagonal entries of $\mathbf{R}$ are monotonically decreasing and the diagonal elements of $R$ are the largest elements in the row. 


\begin{theorem}
If \textbf{B} is rank deficient with rank $M-\ell$, \textbf{R} can be decomposed as \cite{stewart1984rank,chan1987rank}:

\begin{equation}
    \mathbf{R}=\begin{bmatrix}
    \mathbf{R_{11}} & \mathbf{R_{12}} \\ \mathbf{0} & \mathbf{R_{22}},
    \end{bmatrix}
\end{equation}
where $\mathbf{R_{22}}$ is $\ell\times \ell$ and $\sigma_{M-\ell+1}(\mathbf{B})\leq||\mathbf{R_{22}}||_2$, where $\sigma_i(\mathbf{B})$ denotes the $i^{th}$ singular value of \textbf{B} and $||\cdot||_2$ denotes the 2-norm. 
\end{theorem}

\textit{Proof:} See Ref. \cite{chan1987rank}.

Therefore, if $||\mathbf{R_{22}}||_2$ is small, \textbf{B} has at least $\ell$ small singular values and the rank of $\textbf{B}$ can be estimated as $M-\ell$. 

We now assume that $||R_{22}||_2$ is small. In this case, the first $M-\ell$ columns of $Q$ span the column space of $\textbf{B}$ (which therefore has rank $M-\ell$). By construction, the first $M-\ell$ columns of $\mathbf{B\Pi}$ have rank $M-\ell$ and therefore span the column space of $\textbf{B}$. Therefore, one must only retain the first $M-\ell$ columns of $\mathbf{B\Pi}$; the rest are redundant.

Importantly, the RRQR algorithm can also be useful when one would like to obtain a ranking of the best (most linearly independent) columns. The permutation matrix $\mathbf{\Pi}$ can be viewed as a ranking of the ``most linearly independent'' columns of a matrix $\mathbf{\Omega^{(2)}}$. One can then retain the $M_{red}$ ``best'' columns to form $\mathbf{\Omega^{QR}}(M_{red})$. 

\subsection{RRQR for selecting time-shifts}

The RRQR algorithm can be applied to the problem of the selection of node/time-shift combinations. Each column of the time-shifted reservoir matrix $\mathbf{\Omega^{(2)}}$ refers to a specific node/time-shift combination. To determine the set of rank-optimal $M_{red}$ nodes and time-shifts, the QR-factorization with column pivoting is performed on $\mathbf{\Omega^{(2)}}$ and the highest ranked $M_{red}$ combinations (read off from $\mathbf{\Pi}$) are retained.

A significant benefit of this method is that it does not require a model of the system, and therefore is directly applicable to analog RCs, for which a model may not exist. Additionally, the method is not task dependent, as it relies on maximizing the space spanned by the retained columns of the reservoir matrix instead of minimizing the training error for a specific task. In the following sections, we will demonstrate that the same choice of time-shifts obtained using rank-optimization results in significantly improved performance (over randomly-selected time-shifts) for both the prediction and observer tasks for a given chaotic system for both opto-electronic and $tanh$-based reservoir computers

\section{RRQR time-shift rank-optimization for an opto-electronic delay reservoir}
The concept of delay-based reservoir computing was first developed in 2011 \cite{appeltant2011information} and was applied to opto-electronic systems with delayed feedback shortly afterwards \cite{paquot2012optoelectronic,larger2012photonic,soriano2013optoelectronic,chembo2020machine}. These systems have shown promise for high-speed computations with small size and low power consumption and have demonstrated utility for RF demodulation \cite{dai2021classification}, time series prediction \cite{soriano2013optoelectronic,larger2012photonic}, nonlinear channel equalization \cite{antonik2016online,duport2016virtualization,ortin2015unified,paquot2012optoelectronic}, and packet header recognition \cite{qin2017numerical}.

\begin{figure}
\centering
\includegraphics[width=0.48\textwidth]{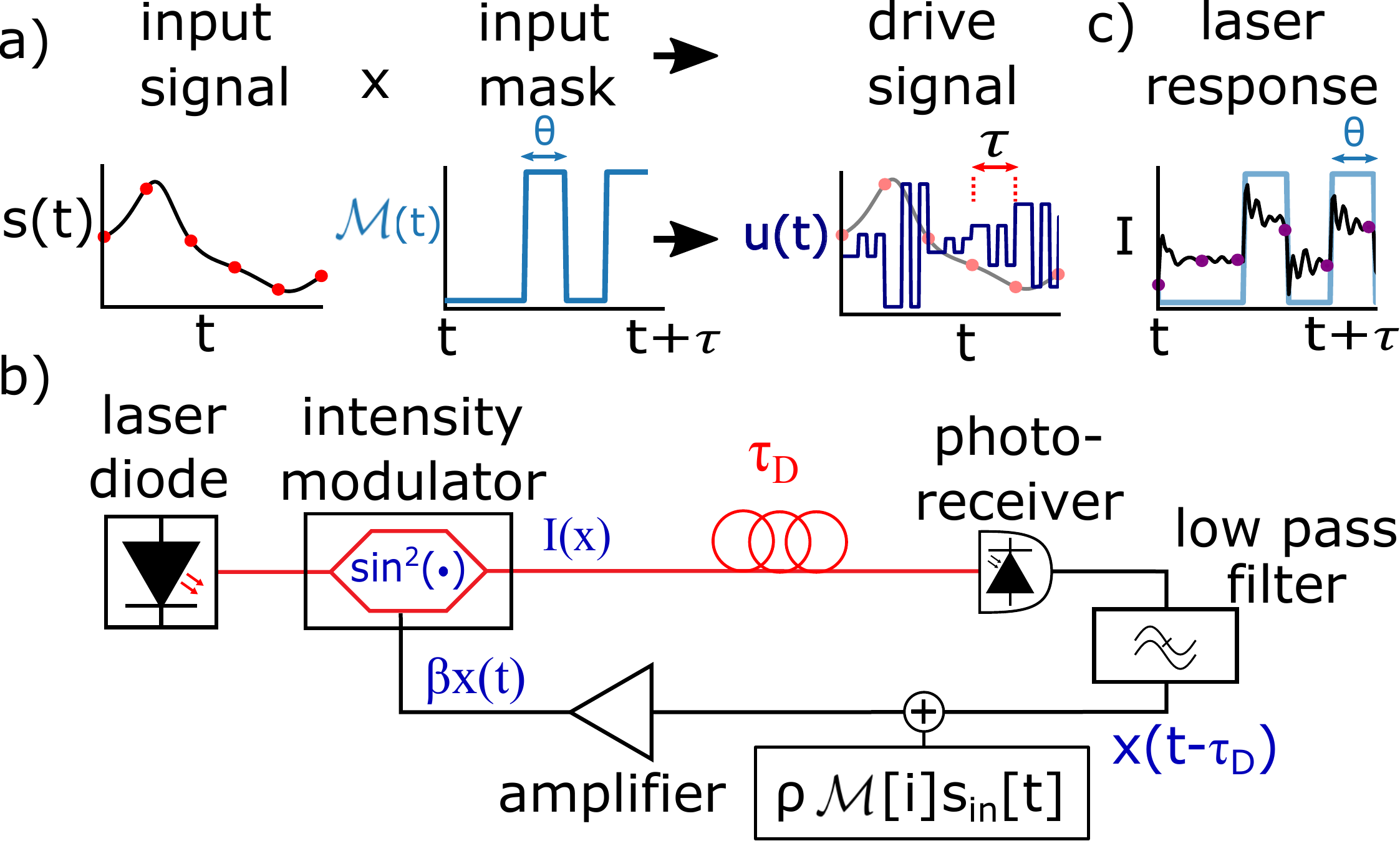}
\caption{Illustration of delay-based opto-electronic reservoir computer. (\textit{a}) Forming the drive signal from the input signal and the input mask {\color{cerulean}$\mathcal{M}(t)$}. (\textit{b}) Opto-electronic RC schematic. $\mathcal{I}$ represents the optical intensity transmitted through the intensity modulator. (\textit{c}) Illustration of the opto-electronic oscillator response (black) with the mask (blue) and sampled $\chi$ values (purple). }
\end{figure}

In an opto-electronic RC, illustrated in Fig. 1b, the output of a fiber-coupled continuous wave laser is sent through an intensity modulator, which provides a sinusoidal nonlinearity. The light that is transmitted through the modulator travels through a fiber delay line before being detected by a photoreceiver. The electrical signal from the photoreceiver is combined with the input signal, amplified, and applied to the RF port of the intensity modulator, completing the feedback loop. This opto-electronic feedback loop forms a dynamical system that is used as a reservoir, where the ``virtual'' nodes come from time multiplexing and the space-time interpretation of delay systems \cite{arecchi1992two,appeltant2011information,yanchuk2017spatio,hart2019delayed}. 

The input signal must be pre-processed in time-multiplexed reservoir systems. The procedure is depicted in Fig. 1a. A periodic input mask $\mathcal{M}(t)$ of period $\tau_{in}$ is applied to the sampled input data (red dots) to form the drive signal (dark blue); $\tau_{in}$ does not have to be equal to the delay time $\tau_d$ \cite{stelzer2020performance,hulser2022role}, though here we choose $\tau_{in}=\tau_d$, as was done in Refs. \cite{carroll2021adding,van2017advances,larger2012photonic,sugano2019reservoir,harkhoe2020,appeltant2011information,larger2017high,soriano2013optoelectronic,dai2021classification}. 
The input mask is analogous to $\mathbf{W^{in}}$ in that it provides a different weighting of the input signal to each time-multiplexed node. Indeed, any $\mathbf{W^{in}}$ can be represented by the input mask $\mathcal{M}(t)$ by choosing $\mathcal{M}(t)=\mathbf{W^{in}_k}$ with $k=\mod(\lfloor t/\theta \rfloor,M)$. 
When one period of the mask is complete, the mask is then applied to the next value of the sampled input signal. Therefore, the reservoir updates at a rate of one input sample per unit $\tau_{in}$. As depicted in Fig. 1a, the input mask is piece-wise constant with duration $\theta$, referred to as the node time. In this work 40\% of the mask values are chosen randomly from the uniform distribution between -1 and +1, and the remaining elements are set to zero. We found this type of mask to give the best results, as discussed in Section \ref{maskopt}.

The photoreceiver output voltage is detected and sampled at a rate $1/\theta$, as depicted in Fig. 1c, to obtain the reservoir node states. The voltage is shown as a black line, with purple dots indicating the sampled values used as the reservoir output node states. 

The opto-electronic RC can be modeled by the following delay differential equation \cite{larger2012photonic}:

\begin{equation}
\label{eq:OEO}
\tau_L \dot{v}(t) = -v(t) + \beta \sin^2\bigg(v(t-\tau_d) + \phi + \rho \mathcal{M}(t)s(t)\bigg),
\end{equation}
where $v$ is proportional to the voltage applied to the intensity modulator, $\tau_L=4\theta$ is the low-pass filter time constant, $\beta=0.8$ is the round trip gain, $\tau_d=M\theta$ is the round trip delay time, and $\phi=0.2$ is the modulator DC bias phase. The system is driven by $\rho \mathcal{M}(t)s(t)$, where $\rho=0.4$ is a scale factor, and $s(t)$ is the continuous-time input signal. $s(t)$ is obtained from the discrete-time, sampled input signal $s_d[k]$ by $s(t)=s_d[\lfloor t/\tau_{in} \rfloor]$. 
In this work, $\theta=40$, and Eq. 4 is integrated using Heun's method with a time step of 1.

\subsection{Optimizing the Input Mask}
\label{maskopt}

An important question for opto-electronic reservoir computer design is how to choose the input mask. In this section, we investigate this question. It seems likely that driving all the virtual nodes of the reservoir with the input signal could decrease the diversity of the reservoir response; we can quantify this diversity by varying the fraction of virtual nodes that are driven and measuring the resulting entropy of the reservoir. We find that there is an optimal value of $f_W$, in this case 40\%, that results from a tradeoff between entropy and covariance rank.

\subsubsection{Estimating Entropy}
Measuring entropy requires a partitioning of the dynamical system. Reference \cite{xiong2017} lists a number of ways to do this partitioning, although different partitions can give different results for the entropy.  It was found that the permutation entropy method \cite{bandt2002} avoided this coarse graining problem because it creates partitions based on the time ordering of the signals.  Each individual node time series $\chi_i(t)$  was divided into windows of 4 points, and the points within the window were sorted to establish their order; for example, if the points within a window were 0.1, 0.3, -0.1 0.2, the ordering would be 2,4,1,3. Each possible ordering of points in a signal $\chi_i(t)$ represented a symbol $\psi_i(t)$.
  
The method of \cite{bandt2002} was adapted for a reservoir computer in  \cite{carroll2021b}. At each time step $t$, the individual node signals were combined into a reservoir computer symbol $\Lambda(t)=[\psi_1(t), \psi_2(t), \ldots \psi_M(t)]$. With $M=100$ nodes there were potentially a huge number of possible symbols, but the nodes were all driven by a common drive signal, so only a tiny fraction of the symbol space was actually occupied, on the order of tens of symbols for the entire reservoir computer.

If $K$ total symbols were observed for the reservoir computer for the entire time series, then the reservoir computer entropy was
\begin{equation}
\label{entropy}
H =  - \sum\limits_{k = 1}^K {p\left( {{\Lambda _k}} \right)\log \left( {p\left( {{\Lambda _k}} \right)} \right)} 
\end{equation}
where $p(\Lambda_k)$ is the probability of the $k^{th}$ symbol.

\subsubsection{Entropy vs. Error}
For this simulation, the elements of the input mask ${\bf W}^{in}$ were chosen from a random distribution between -1 and 1, but only a fraction $f_W$ of the elements were nonzero. For each value of $f_W$, 20 realizations  of ${\bf W}^{in}$ were created and the numbers that are plotted are the average from these 20 realizations. 

\begin{figure}
\centering
\includegraphics[scale=0.8]{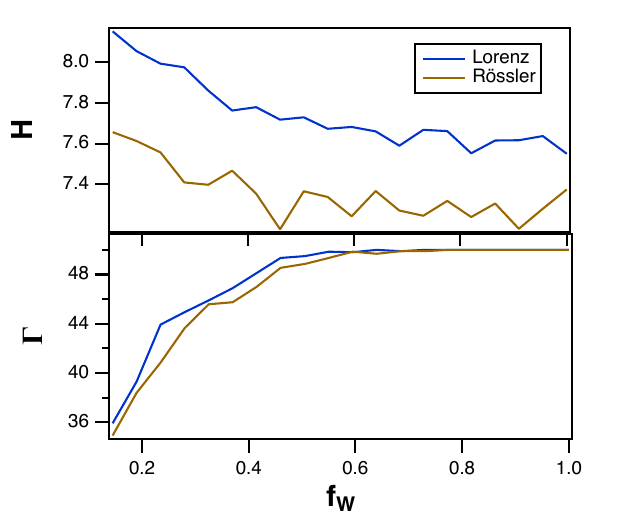}
\caption{\label{oeoentropy} The top plot shows the entropy $H$ of the optoelectronic reservoir driven by the Lorenz or the R{\" o}ssler $x$ signal as the fraction of nonzero entries in the input mask $f_W$ varied. The bottom plot shows the covariance rank $\Gamma$ of the reservoir with the two different input signals}
\end{figure}

Figure \ref{oeoentropy} shows the entropy $H$ as a function of the input fraction $f_W$ when the reservoir was driven with the Lorenz or the R{\" o}ssler $x$ signal.

Decreasing the fraction of virtual nodes directly connected to the input signal increases the entropy of the optoelectronic reservoir. Figure \ref{oeoentropy} also shows the covariance rank of the reservoir driven by the two different input signals.

\begin{figure}
\centering
\includegraphics[scale=0.8]{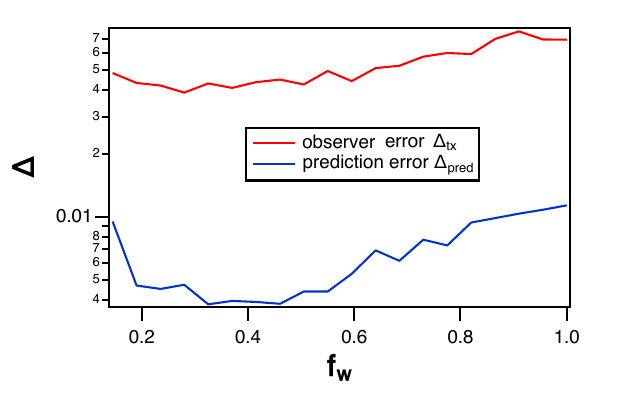}
\caption{\label{oeolortest} Observer and prediction testing errors for the optoelectronic reservoir driven by the Lorenz $x$ signal as the fraction of nonzero input mask elements $f_W$ varies.}
\end{figure}
Figure \ref{oeolortest} shows that the observer and prediction testing errors both go through a minimum as $f_W$ varies. Figure \ref{oeoentropy} explains the reason for this minimum; as $f_W$ gets smaller the reservoir entropy $H$ gets larger, and larger entropy should lead to smaller fitting errors, but the reservoir covariance rank $\Gamma$ gets smaller. Smaller covariance rank should lead to larger fitting errors, so the two effects lead to a minimum in the testing and prediction errors near $f_W=0.4$.

    \begin{figure}[ht!]
    \centering
\includegraphics[width=0.48\textwidth]{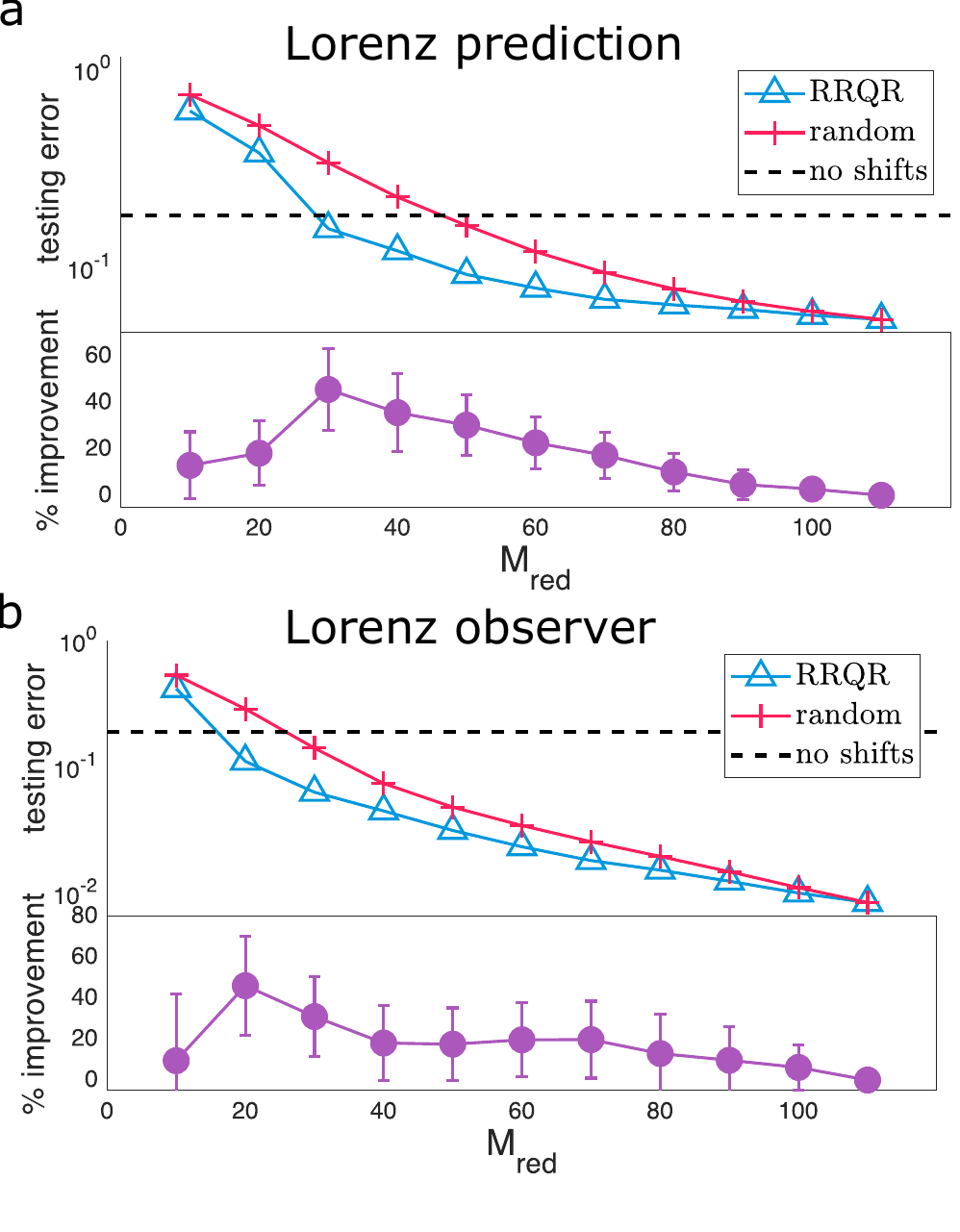}
    \caption{ Comparison of testing NRMSE for the opto-electronic reservoir computer using RRQR-optimized time-shifts and randomly selected time-shifts as a function of $M_{red}$ for the Lorenz tasks with 10 nodes and $\tau_{max}=10$. (\textit{a}) One-step prediction. (\textit{b}) Observer task.}
    \label{fig:lorenz}
\end{figure}

\subsection{Results}

\begin{figure}[!t]
    \centering
\includegraphics[width=0.48\textwidth]{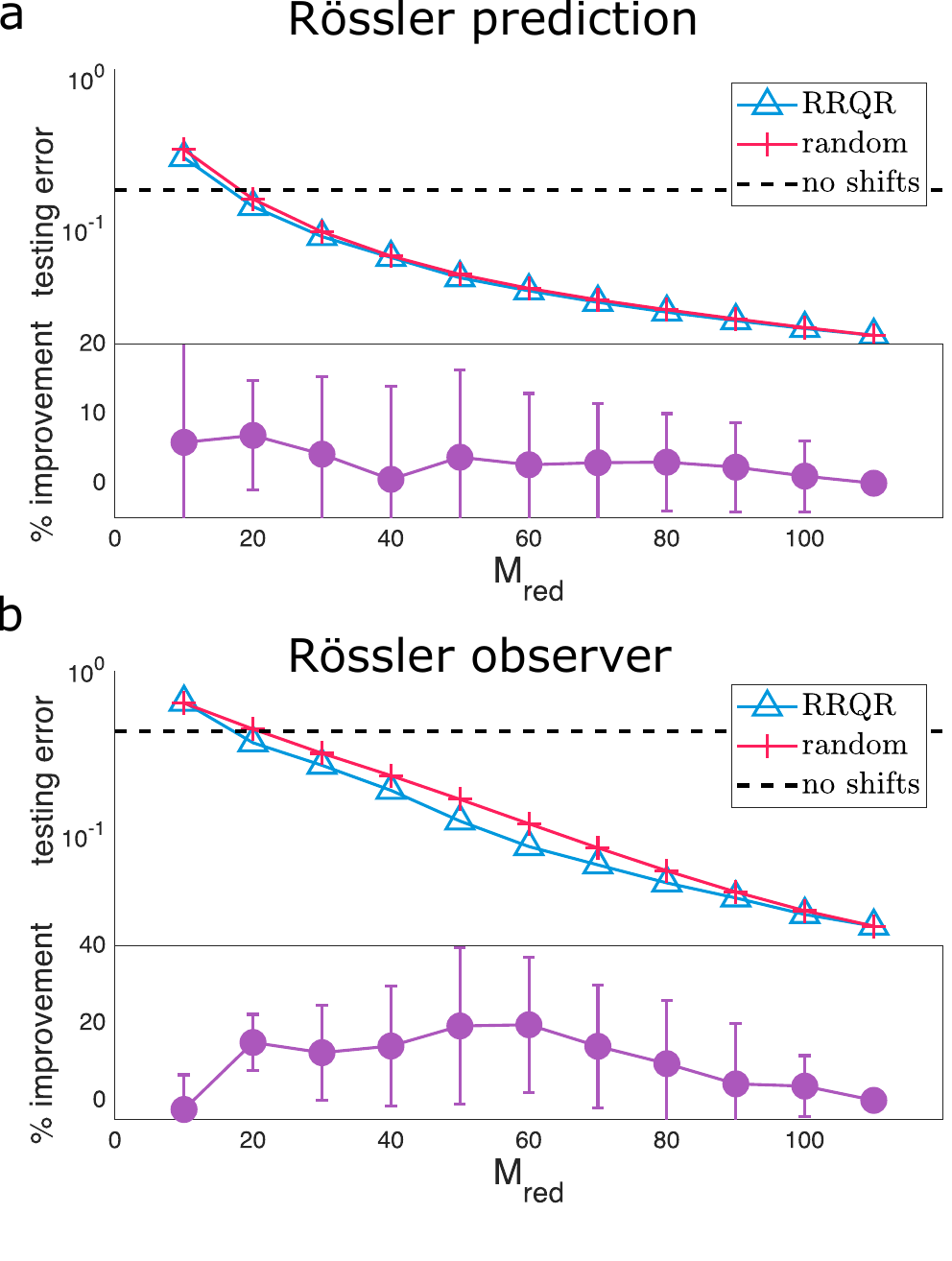}
    \caption{ Comparison of testing NRMSE for the opto-electronic reservoir computer using RRQR-optimized time-shifts and randomly selected time-shifts as a function of $M_{red}$ for the R{\"o}ssler tasks with 10 nodes and $\tau_{max}=10$. (\textit{a}) One-step prediction. (\textit{b}) Observer task.}
    \label{fig:rossler}
\end{figure}

We investigated the effectiveness of the RRQR time-shift rank-optimization described in Section \ref{sec:rrqr} on an opto-electronic RC with $M=10$ nodes performing the prediction and observer tasks on the Lorenz and R{\"o}ssler chaotic systems. The testing NRMSE obtained using the best $M_{red}$ node/time-shift combinations are shown as blue triangles in Fig. \ref{fig:lorenz} (Lorenz) and Fig. \ref{fig:rossler} (R{\"o}ssler). These results are averaged over 20 different, randomly selected input masks.

For comparison, we show the NRMSE obtained using $M_{red}$ randomly selected columns of $\mathbf{\Omega^{(2)}}$ as red crosses in Figs. \ref{fig:lorenz} and \ref{fig:rossler}. This NRMSE is averaged over 20 different, randomly generated input masks and 20 different, randomly selected node/time-shift combinations. We also show the NRMSE obtained using the non-time-shifted reservoir ($\mathbf{\Omega}$) as a black dotted line. 

In all cases, the accuracy improves as $M_{red}$ increases. This is expected, since having a larger state matrix allows for the possibility of a larger state matrix rank. However, the random and rank-optimal state matrices have the same state matrix size for a given $M_{red}$, yet the rank-optimal state matrix generally provides better performance, implying that choosing the correct time-shifts can result in a significant improvement in performance.

For the Lorenz system (Fig. \ref{fig:lorenz}), we find that the testing error using rank-optimal node/time-shift combinations is always better than randomly selected combinations. To quantify this, we compute the percent improvement

\begin{equation}
    \mathrm{\% \; improvement} = 100\times\frac{\Delta_{test}^{rand} - \Delta_{test}^{RRQR}}{\Delta_{test}^{rand}},
\end{equation}
where $\Delta_{test}^{RRQR}$ and $\Delta_{test}^{rand}$ are the testing NRMSE for the rank-optimal and randomly selected node/time-shift combinations, respectively. We see that, for the Lorenz system, the percent improvement provided by the RRQR algorithm is as high as 55\% for the prediction task and 68\% for the observer task. As expected, as $M_{red}$ approaches $M(\tau_{max}+1),$ the testing errors from the rank-optimal and random selections converge, since in both cases all node/time-shift combinations are being used (i.e., the full $\mathbf{\Omega^{(2)}}$ is used). The error bars for the improvement metric were determined by computing the improvement metric for each reservoir realization, and using the standard deviation of this data set as the error bars. We did not compute error bars for the testing errors because any error bar computation for the testing errors for the randomly chosen time-shifts will confuse variation in performance from reservoir to reservoir with variation in performance due to the different randomly selected time-shifts.

For the R{\"o}ssler system (Fig. \ref{fig:rossler}), we similarly find that the testing error using rank-optimal node/time-shift combinations is nearly always better than randomly selected combinations, but the improvement is more modest, especially for the prediction task. We suspect that this performance reduction is due to the long correlation time of the R{\"o}ssler system, which means that the time-shifted signals are not so different from one another. This is supported by Fig. 2, in which the entropy statistic shows that there is less diversity among the different columns when the reservoir is driven by the R{\"o}ssler $x$ variable. In this case, one would not expect the particular selection of node/time-shift combinations to make as much difference as for the Lorenz system which has a much shorter correlation time and for which the reservoir displays a larger entropy.

 \begin{figure}[ht!]
\centering
\includegraphics[scale=0.7]{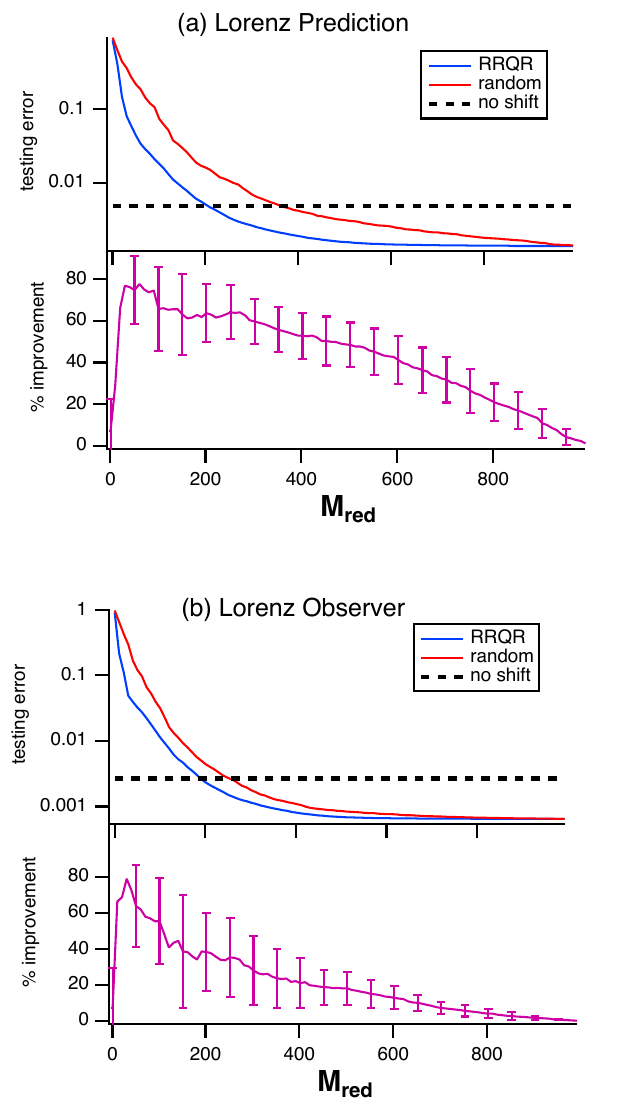}
\caption{\label{tanhlor} Comparison of testing NRMSE for the leaky $tanh$ reservoir computer using RRQR-optimized time shifts and randomly selected time shifts as a function of $M_{red}$ for the Lorenz tasks with 100 nodes and $\tau_{max}=10$. (a) One step prediction of the Lorenz $x$ variable. (b) Observer task of the Lorenz $z$ variable based on the Lorenz $x$ variable.
}
\end{figure}

It is important to note that the RRQR algorithm selects the most linearly independent columns of the full reservoir matrix $\mathbf{\Omega^{(2)}}$; it does not provide an optimal $\tau_{max}$. This explains why both the rank-optimized and randomly-selected time-shifts perform worse than the unshifted reservoir for $M_{red}=10$: The time-shifts retained include large shifts, which, while more linearly independent, are less correlated with the current system state than the unshifted nodes. For small $M_{red}$, the RRQR method performs better than the unshifted reservoir only for small $\tau_{max}$ (not shown). 

We have tested our RRQR method on larger opto-electronic RCs (up to $M=800$) and using different $\tau_{max}$, and found similar results. 

\section{Time-shift rank-optimization for the Leaky Tanh Reservoir}
In the previous section, we showed that rank-optimization significantly improves the performance of an optoelectronic reservoir. In order to demonstrate the generality of this technique, in this section we show that the rank-optimization is also useful for a leaky $tanh$ reservoir \cite{jaeger2001}. This reservoir computer is described by

\begin{equation}
\label{tanhnode}
\begin{split}
{\bf {\chi}}[{n+1}]=&\left(1-\alpha\right)\mathbf{\chi}[n]+\\ &\alpha\tanh\left({{\bf {A}\chi[n]}+\mathbf{W}^{in}s[n]}+1\right)
\end{split}
\end{equation}
 For these simulations, half the elements of ${\bf A}$ were chosen from a uniform random distribution between -1 and 1 while the others were zero, and then the the diagonal of ${\bf A}$ was set to zero.  ${\bf A}$ was then renormalized so its spectral radius was 0.5. The parameter $\alpha=0.35$ was found to minimize the testing error. The signals $\chi_j[i]$ were arranged in a matrix $\Omega$ in the same manner as in Eq. (\ref{omegaeq}) and the reservoir computer was used to fit training and testing signals as in Section \ref{rescomp}, with the testing error being calculated as in Eq. (\ref{eq:training_error}). As in the optoelectronic system, optimization revealed that the best testing error was found when 40\% of the elements of ${\bf W^{in}}$ were nonzero.

 \subsection{Results}
\begin{figure}[t!]
\centering
\includegraphics[scale=0.7]{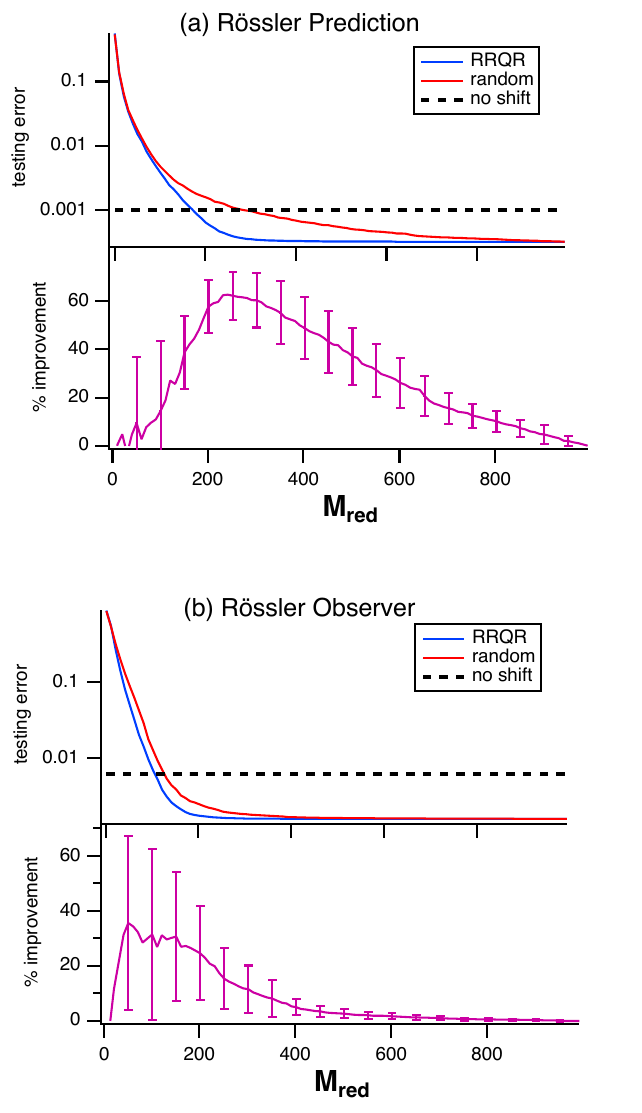}
\caption{\label{tanhross} Comparison of testing NRMSE for the leaky $tanh$ reservoir computer using RRQR-optimized time shifts and randomly selected time shifts as a function of $M_{red}$ for the R{\"o}ssler tasks with 100 nodes and $\tau_{max}=10$. (a) One step prediction of the R{\"o}ssler $x$ variable. (b) Observer task of the R{\"o}ssler $z$ variable based on the R{\"o}ssler $x$ variable.
}
\end{figure}

The testing NRMSE using the rank-optimal $M_{red}$ node/time shift combinations for the leaky $tanh$ reservoir and the Lorenz system is shown in Fig. \ref{tanhlor} and for the R{\"o}ssler system in Fig. \ref{tanhross}.

The improvement in testing error in the leaky $tanh$ reservoir computer gained by using rank-optimal time shift/node combinations is even larger than in the opto-electronic reservoir computer. The opto-electronic system does have more symmetry because the coupling between virtual nodes is a one way ring \cite{hart2019delayed}, while the coupling in the leaky $tanh$ system is random. Similarly to the opto-electronic system, the improvement is larger for the Lorenz tasks than for the R{\"o}ssler tasks.

\section{Conclusions}
We have proposed a technique for selecting the rank-optimal combination of time-shifts for a reservoir computer based on a rank-revealing QR algorithm. We demonstrated that a reservoir computer with our rank-optimized time-shifts displays significantly higher accuracy than with random time-shifts. We have verified that this technique is general by demonstrating improved performance on both analog and digital reservoir computers. Our technique is task-independent and does not rely on a model of the reservoir computer, so it can be straightforwardly applied to both digital and analog reservoir computer implementations.

\textbf{Author Declarations}: The authors have no conflicts of interest to disclose.

\textbf{Data Availability}: The data that support the findings of this study are available from the corresponding author upon reasonable request.

\textbf{Acknowledgements:} JDH and TLC acknowledge support from the Naval Research Laboratory's Basic Research Program and from the Office of the Secretary of Defense through the Applied Research for Advancement of S\&T Priorities (ARAP) program under the Neuropipe project.

%

\end{document}